# On Considering Uncertainty and Alternatives in Low-Level Vision

**Steven M. LaValle** **Seth A. Hutchinson**
Dept. of Electrical and Computer Engineering
and
The Beckman Institute
University of Illinois, Urbana, IL 61801

## Abstract

In this paper we address the uncertainty issues involved in the low-level vision task of image segmentation. Researchers in computer vision have worked extensively on this problem, in which the goal is to partition (or segment) an image into regions that are homogeneous or uniform in some sense. This segmentation is often utilized by some higher-level process, such as an object recognition system. We show that by considering uncertainty in a Bayesian formalism, we can use statistical image models to build an approximate representation of a probability distribution over a space of alternative segmentations. We give detailed descriptions of the various levels of uncertainty associated with this problem, discuss the interaction of prior and posterior distributions, and provide the operations for constructing this representation.

## 1 INTRODUCTION

Image segmentation has been an active topic in low-level vision research for over two decades. The segmentation problem is typically defined as the task of extracting a set of homogeneous regions (called segments) from an image, often for the purpose of higher-level processing. Although considerable research effort has yielded a number of approaches to the problem, segmentation remains a difficult problem in its general formulation. As Horn has pointed out, one of the primary difficulties in evaluating a segmentation method is the lack of a clear definition of the "correct" segmentation [8]. This definition usually depends on the intended application of the segmentation result. Szeliski argues that low-level image models often underconstrain the solution, and advocates the use of uncertainty estimation [23]. Jain and Binford assert that a key problem with vision research is that researchers either assume there is no segmentation problem or assume that it has been solved [10].

In light of these observations we do not consider segmentation as a single isolated problem with an optimal solution that is yet to be discovered, which is the approach taken in traditional statistical segmentation paradigms, such as statistical clustering [11, 22, 24], Markov random fields [7, 14, 23], and probabilistic relaxation [17, 18]. Instead we use statistical segmentation models to determine a Bayesian posterior distribution over a set of alternative image partitions. Some higher-level vision approach which requires a segmentation as a premise can utilize alternative segments and segmentations and/or make inferences about effectiveness of the statistical models based on the properties of the probability distribution over alternatives. These concepts and arguments will be elucidated shortly.

For a precise formulation of segmentation, it is useful to consider the description given by Horowitz and Pavlidis [9]. Intuitively, one wants the segments in a segmentation to be homogeneous or uniform in some sense. Abstractly, this can be formalized with a ***homogeneity predicate*** (or ***uniformity predicate***). Consider an image, $D$, as a set of elements, with each element containing some information. It is assumed that we are given some homogeneity predicate, $H$, which applies to subsets of $D$. If $R$ is some subset of $D$, then the predicate, $H$, returns ***true*** if the set is homogeneous or ***false*** otherwise.[1]

A segmentation is defined for some $H$ and $D$ as a collection of disjoint nonempty subsets $X_1, \ldots, X_n$ such that

1. $\bigcup X_i = D$
2. $X_i$ is connected (optional and must be consistently defined)
3. $H(X_i) = \textbf{true} \;\; \forall X_i$
4. $H(X_i \cup X_j) = \textbf{false} \;\; \forall X_i \neq X_j$ .

We consider a segment to be a maximal region such that $H(R) = \textbf{true}$.[2]

We represent the homogeneity predicate using a ***parameter space***. The parameter space directly captures the notion of homogeneity: every region has a parameter value (a point in the parameter space) associated with it, which is unknown to the observer.[3] Two regions are defined to be homogeneous if they share the same parameter value. As an example the parameter space could represent the coefficients of a polynomial surface describing image intensities, as in the model used by Silverman and Cooper [22]. If the parameter value is known for each region, an ideal segment is a maximal set of connected regions sharing the same parameter value. It is assumed that the observer (receiving information only from the image) does not know the associated parameter value for any of the regions. If it is possible to determine the parameter values for each of the regions, then the ideal segmentation can be trivially determined.

[1] This is primarily a conceptual formulation, and in practice a logical predicate is usually not specified in this form.

[2] In Section 2, we will introduce alternative, but equivalent definitions of segments and segmentations.

[3] The observer refers to the machine, which receives only the image data.

Although the definition of homogeneity is straightforward to express, the problem of determining maximally homogeneous image subsets is quite challenging. This is due primarily to noise that occurs in an imaging process. Although models can be formulated that accurately describe the expected relationships between the image data, the added noise process forces the need for considering uncertainty.

The space of alternatives is therefore often underconstrained when using low-level models [23], and one approach is to introduce more constraints through the use of higher-level models, for instance, at the recognition level. For this to occur, it is unreasonable to select a single, apparently best, segmentation to send to the higher-level process. The single segmentation has been formed by making all of the decisions using low-level models, and all other information is lost. For the higher-level models to participate in the segmentation process, it seems useful to at least give some set of alternative segmentations. Additional evidence can then begin to be applied by the higher-level process to constrain the space of segmentations, eventually resulting in a unique solution.

Rather than simply representing a set of alternatives, consider also obtaining probabilities for each of the alternatives. The probabilities give much more information than is present in the set of alternatives alone. For instance, if the leading segmentation obtains ten times the probability of its leading competitor, then the confidence in the segmentation should be high. If the top ten segmentations have approximately the same probability, some other process may have to be performed to further constrain the solution.

For a typical application, it is useful to know the degree to which a particular image model is providing information regarding the segmentation. With a probability distribution over segments and segmentations available, a formal measure of information content can be directly quantified. One natural measure is the information entropy, which is a function of a probability distribution [2]. Szeliski argues that a measure of uncertainty can be used to guide search, indicate when more sensing is required, and integrate new information [23].

To build the probability distributions using statistical models, we have developed a Bayesian formalism. The structure of computation resembles a regression tree [5, 6] or a taxonomic hierarchy [16]. Hence, at least structurally, our decomposition of the problem can be considered as a special case of Bayesian networks, which have been applied to other computer vision problems. Agosta, and Binford et al. have considered them for model-based object recognition applications [1],[3]. Sarkar and Boyer have proposed Bayesian networks for a hierarchical organization of perceptual features [20].

The Bayesian formalism also provides a natural way to combine evidence from several models. In general, a Bayesian approach begins with some prior distribution and some evidence, and yields a posterior distribution. A multiple model approach could treat the posterior distribution from one model as the prior distribution for a second (possibly higher-level) model. The second posterior distribution reflects the application of both models. This concept can be applied directly to segment and segmentation distributions, and also to region pairs, as discussed in Section 5.

As shown in Figure 1, we consider four levels of uncertainty associated with the segmentation problem. The paper is organized around this decomposition of the uncertainty. Section 2 formally defines the regions, segments, and segmentations; however, we briefly describe their relationships here. A region is a (typically small) subset of the image data that is assumed to be homogeneous. A segment is a set of regions that together are hypothesized as a maximally homogeneous subset in the image. A segmentation is a set of segments that covers all the data in the image, representing a partition of $D$.

We describe the lowest level as data-level uncertainty, which is modeled by a statistical description of the noise that occurs in the imaging process. Models of this form have often been considered for segmentation, as in [4, 22]. In Section 5 we describe how models of this type can be used to make probabilistic assessments about homogeneity at the region level. Section 3 describes how the probabilistic assessments at the region level are used to construct a probability distribution over alternative segments, representing segment-level uncertainty. Section 4 discusses how the uncertainty obtained at the segment level is used to construct a probability distribution over a set of segmentations. Although we would prefer to cover the levels in a top-down manner, the dependencies between the concepts make the discussion of segment-level uncertainty a more natural place to begin.

Section 7 briefly discusses the algorithm issues involved in efficiently building the probability distributions. An experimental example is presented in Section 8. Detailed descriptions of our algorithms and several dozen experiments are given in [12]. Some general conclusions are presented in Section 9.

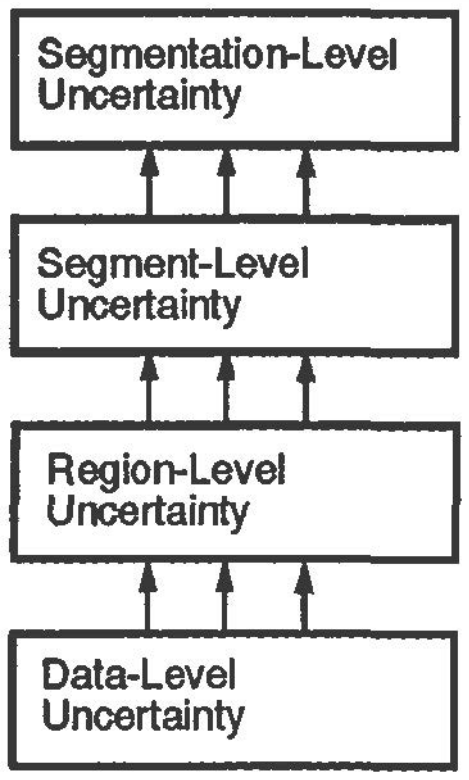

Figure 1: A schematic representation of the different levels of uncertainty associated with segmentation

## 2 REGIONS, SEGMENTS, AND SEGMENTATIONS

Recall that the input to the segmentation system is a set $D$ of points. Associated with each element $D[i,j]$ is the representation of the point, which may be an intensity value and/or a set of coordinates in $\Re^3$, or may be other image information. The elements of $D$ are typically arranged in an array, and adjacencies can be considered in a standard way. A given point $D[i,j]$ has a set of points called *neighbors* to which it is *adjacent*. Using standard four-neighbors, this set is: $D[i-1,j]$, $D[i+1,j]$, $D[i,j-1]$, $D[i,j+1]$. One could also consider eight-neighbors by considering diagonally related points as adjacent.

A *region*, $R$, is some connected subset of $D$. By *connected* we mean that for any $D[i_1,j_1]$, $D[i_2,j_2] \in R$, there exists some sequence of elements of $R$, $\langle D[k_m,l_m]\rangle_{m=1,\ldots,n}$, in which each $D[k_{m-1},l_{m-1}]$ is adjacent to $D[k_m,l_m]$, and $D[k_1,l_1] = D[i_1,j_1]$, $D[k_n,l_n] = D[i_2,j_2]$. Two regions, $R_1$ and $R_2$, will be called *adjacent* if there exist some $D[i_1,j_1] \in R_1$ and $D[i_2,j_2] \in R_2$ that are adjacent.

For a given problem, we work with a pairwise-disjoint set of regions, $\mathcal{R}$, in which every element of $D$ is contained in some region. It is often profitable to begin with some initial partition of the image into small regions, and construct new segmentations by combining regions. This the standard approach taken in the region merging paradigm [19]. One justification for this is the savings in complexity achieved by considering this smaller set of possible segmentations. Another reason is that often some minimal number of points is required in a region before the statistical models can be effectively employed. The initial segmentation represents the starting point in a region-merging algorithm. For instance, Silverman and Cooper begin with an initial image of *blocks*, which corresponds to an initial partition of the image into a grid of square regions [22]. Blocks are merged to yield *clusters*, which correspond to segments. The implication of starting with $\mathcal{R}$ is that there are many *image* partitions that are not considered.

A *segment*, $T$, is a connected set of regions (e.g., $T = \{R_1, R_2, R_3\}$ is a segment consisting of three regions). A set of regions is *connected* if their union is connected, in the sense as defined above.

A *segmentation*, $S$, denotes a set of segments that forms a partition of $\mathcal{R}$. Note that a segmentation implicitly defines a partition of $D$.

## 3 SEGMENT-LEVEL UNCERTAINTY

We will construct an approximate representation of the segmentation sample space through the consideration of a number of approximate representations of individual segment sample spaces, each of which corresponds to a set of hypotheses about individual segments in the image. In this section, we define the segment sample space, and describe how approximate segment sample space representations are constructed. In Section 4, we show how approximate segmentation sample space representations are constructed, using these approximate segment sample space representations as a basis.

For some region $R_i \in \mathcal{R}$, let $\Theta_i$ be the set of all possible segments that contain $R_i$. Specifically,

$$\Theta_i = \{T \subseteq \mathcal{R} \mid T \textit{ is connected}, \ R_i \in T\}. \quad (1)$$

Note that $\Theta_i$ always contains at least two elements: the singleton $\{R_i\}$ and the entire set $\mathcal{R}$ (provided $D$ is connected). For any such $\Theta_i$, there is a corresponding segment sample space that describes both $\Theta_i$ and the probabilities associated with each subset of $\Theta_i$. Specifically, a segment sample space is defined as $\mathcal{T}_i = (\Theta_i, \mathcal{B}_i, P_i)$ in which $\Theta_i$ is defined as in (1), $\mathcal{B}_i$ is the set of all events (i.e., subsets of $\Theta_i$), and $P$ is a probability mapping on $\mathcal{B}_i$. (To simplify notation, we usually omit the subscript on the probability mapping $P_i$).

For real image applications, the set of segments, $\Theta_i$, will be extremely large; the set $\mathcal{B}_i$ is exponentially larger. Therefore, it is infeasible to explicitly enumerate the elements in either $\Theta_i$ or $\mathcal{B}_i$. To deal with these combinatoric issues, we now introduce an implicit representation for elements of $\mathcal{B}_i$, a representation for approximations to $\mathcal{T}_i$, and a mechanism by which any approximation of $\mathcal{T}_i$ can be refined to yield a more accurate approximation.

Each element of $B \in \mathcal{B}_i$ corresponds to a set of segments. We can uniquely identify this set by specifying: (i) the set of all regions common to every segment in $B$, and (ii) a particular set of regions not included in any segment in $B$. Specifically, the ***inclusion set***, $I$, is the set of regions common to every segment in $B$ (note that $I$ always includes $R_i$). The ***exclusion set***, $E$, is a set of regions that are not included in any segment in $B$. To eliminate redundant representations, we require each element of $E$ to be adjacent to some region in $I$. Note that $I \cap E = \emptyset$. Using this notation, we define $\tau(I,E)$, which maps to some $B \in \mathcal{B}_i$, as

$$\tau(I,E) = \{T \in \Theta_i : I \subseteq T, \ E \cap T = \emptyset\}. \quad (2)$$

Thus, $\tau(I,E)$ specifies the set of all segments that include all regions in $I$, and exclude all regions in $E$. Every event $B \in \mathcal{B}_i$ has a well-defined representation in terms of $I$ and $E$ sets [12].

Given this representation for subsets of $\Theta_i$, we now turn to the construction of approximations of $\mathcal{T}_i$. We will formally define the notions of cover and refinement, and then describe them in terms of Figure 2. We will construct approximations of $\mathcal{T}_i$ that explicitly represent those segments that have high probability values, while only implicitly specifying large subsets of segments that have small probability values. However, for any approximation to $\mathcal{T}_i$, every segment should be represented, either explicitly or implicitly. To this end, we define a ***$\mathcal{T}$-cover***, $C_i$, of $\mathcal{T}_i$ to be a set of pairwise-disjoint elements in $\mathcal{B}_i$ that form a partition of $\Theta_i$.

If the probabilities for the elements in $C_i$ are known, we can consider $C_i$ to represent an approximation of $\mathcal{T}_i$. It is approximate because probabilities are not associated with the singletons in $\mathcal{B}_i$, but only with those elements that are explicitly contained in $C_i$. Since the elements of $C_i$ form a partition of $\Theta_i$, every element of $\Theta_i$ is represented in $C_i$, either explicitly (in the case of singleton events) or implicitly (in the case of non-singleton events).

The notion of quality of approximations can be formalized as a partial ordering on $\mathcal{T}$-covers. This gives rise to a lattice of partitions of $\Theta_i$. Given two $\mathcal{T}$-covers, $C_i$ and $C_i'$, $C_i' \preceq C_i$ if and only if for all $B' \in C_i'$ there exists some $B \in C_i$ such that $B' \subseteq B$. In other words, $C_i' \preceq C_i$ if $C_i'$ can be obtained by partitioning some of the elements of $C_i$.

We denote by $C_i^\omega$ the set of all singleton events in $\mathcal{B}_i$. Thus, $C_i^\omega$ is an exact representation of $\mathcal{T}_i$; all of the elements of $\Theta_i$ are explicitly represented, and the probability for each is given. Hence, in this case the entire probability map is fully determined (since the probability for any $B \in \mathcal{B}_i$ can be obtained by summing the probabilities $P(\{T\})$ for each $T \in B_i$). Thus, $C_i^\omega \preceq C_i$ for all $\mathcal{T}$-covers $C_i$. The poorest approximation of $\mathcal{T}_i$ is $C_i^0 = \{\Theta_i\}$. We know that $P(\Theta_i) = 1$; however, the probabilities of the other events in $\mathcal{B}_i$ cannot be directly determined. Thus, $C_i \preceq C_i^0$ for all $\mathcal{T}$-covers $C_i$.

Procedurally, in order to construct approximations to $\mathcal{T}_i$, we begin with $C_i^0$, and derive a sequence of $\mathcal{T}$-covers, $C_i^k$, such that $C_i^{k+1} \preceq C_i^k$. Each step in this sequence corresponds to a single *$\mathcal{T}$-refinement* operation. Specifically, given a $\mathcal{T}$-cover $C_i^k$, an event, $B_\rho = \tau(I_\rho, E_\rho) \in C_i^k$, and a region $R_\rho \notin I_\rho \cup E_\rho$, we define a new $\mathcal{T}$-cover by

$$C_i^{k+1} = C_i^k + \tau(I_\rho \cup \{R_\rho\}, E_\rho) + \tau(I_\rho, E_\rho \cup \{R_\rho\}) - B_\rho. \quad (3)$$

The expression above is termed the *$\mathcal{T}$-refinement mapping.* The region $R_\rho$ is termed the *$\mathcal{T}$-refinement region.* In order to ensure that only connected sets of regions are represented in the new $\mathcal{T}$-cover, we require the $\mathcal{T}$-refinement region, $R_\rho$ to be adjacent to some region in $I_\rho$. The event $B_\rho$ is termed the *$\mathcal{T}$-refinement event.* The $\mathcal{T}$-cover, $C_i^{k+1}$, is termed the *refined $\mathcal{T}$-cover* with respect to $C_i^k$. The only difference between $C_i^k$ and $C_i^{k+1}$ is the replacement of $B_\rho$ by $\tau(I_\rho \cup \{R_\rho\}, E_\rho)$ and $\tau(I_\rho, E_\rho \cup \{R_\rho\})$. These two new events will be termed *$\mathcal{T}$-refined events.* Thus, the $\mathcal{T}$-refinement operation has the effect of partitioning the event $B_\rho$ into two new subsets of $B_\rho$: the segments in $B_\rho$ that include $R_\rho$ are in $\tau(I_\rho \cup \{R_\rho\}, E_\rho)$ and the remaining elements of $B_\rho$ (all those that exclude $R_\rho$) are in $\tau(I_\rho, E_\rho \cup \{R_\rho\})$. Each singleton in $\mathcal{B}_i$ represents a single segment. We will refer to these events as *$\mathcal{T}$-ground events,* since such events cannot be refined.

Figure 2 provides a pictorial description of covers and refinements. Each of the six long, rectangles abstractly represents a partition of $\Theta_i$ (by dividing the rectangles with vertical bars), and hence a $\mathcal{T}$-cover. From the top down, each cover is obtained by performing a refinement on the previous cover. The shaded regions (one per cover) indicate the $\mathcal{T}$-refinement event, and each of these points to the two $\mathcal{T}$-refined events which occur in the next cover. At the first level, $\Theta_i$ is the first $\mathcal{T}$-refinement event, and the successive covers provide better approximations of $\mathcal{T}$ (assuming probabilities are known for each element in the cover).

# 4 SEGMENTATION-LEVEL UNCERTAINTY

Now that the segment sample space has been defined, its relationship to the segmentation sample space ($\mathcal{S}$) will be discussed. It will turn out that $\mathcal{T}$ can be used as a building block with which representations of events on $\mathcal{S}$ can be constructed. It is necessary to relate the distribution over $\Theta_i$ to the distribution on $\mathcal{S}$ since traditionally one is interested in full image segmentations. This section culminates in the presentation of $\mathcal{S}$-covers and $\mathcal{S}$-refinement, which are analogs of the $\mathcal{T}$ versions.

Let $\Pi$ denote the set of all segmentations that could be constructed from the regions, $\mathcal{R}$. At one extreme, $\Pi$ includes the partition induced by the original regions. At another extreme, $\Pi$ contains the partition corresponding to combining all regions into one segment. In this section we consider $\Pi$ in a probability space, by considering subsets of $\Pi$ and their associated probabilities. Hence a probability, $P(\{S\})$, is associated with each possible $S \in \Pi$. The segmentation sample space is represented by the probability triple $\mathcal{S} = (\Pi, \mathcal{A}, P)$. In the triple, $\mathcal{A}$ represents the set of all subsets of $\Pi$ (i.e., the power set of $\Pi$), and $P$ denotes a probability mapping (distinct from the $\mathcal{T}$ probability mapping), defined on $\mathcal{A}$.

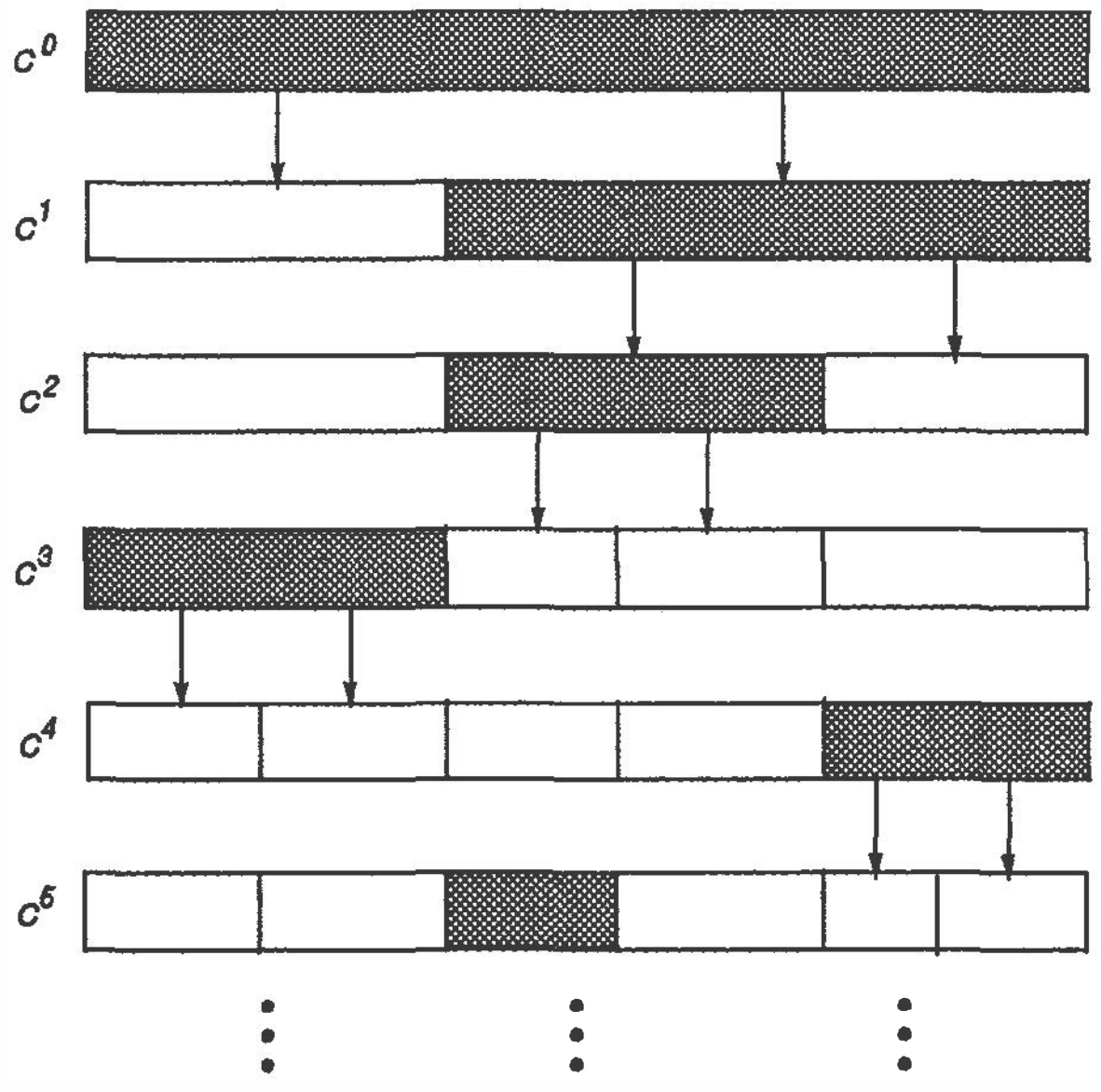

Figure 2: A sequence of covers

## 4.1 The segment-to-segmentation mapping

The relationship between a particular $\mathcal{T}_i$ and $\mathcal{S}$ is specified by the function $f_i : \mathcal{B}_i \rightarrow \mathcal{A}$. For a $\mathcal{T}$-ground event, denoted by $\{T\}$,[4] we define $f_i$ by (see Figure 3)

$$f_i(\{T\}) = \{S \in \Pi : T \in S\}. \quad (4)$$

The event $f_i(\{T\}) \in \mathcal{A}$ is the set of all segmentations that include the segment $T$. Since every $T \in \Theta_i$ contains $R_i$ and segments in a segmentation are disjoint, we have

$$f_i(\{T_1\}) \cap f_i(\{T_2\}) = \emptyset \quad \forall T_1, T_2 \in \Theta_i, \quad T_1 \neq T_2. \quad (5)$$

In other words, no single segmentation can contain two distinct segments that belong to the same $\Theta_i$, since by the definition of $\Theta_i$, such segments would overlap. Using (4) and (5), we define the mapping for a general event, $B \in \mathcal{B}_i$ as

$$f_i(B) = \bigcup_{T \in B} f_i(\{T\}). \quad (6)$$

By applying $f_i$ to each $\mathcal{T}$-ground event of $\mathcal{T}_i$ we obtain a set of events that form a partition of $\Pi$, with each set in the partition corresponding to some segment from $\mathcal{T}_i$.

[4]We use $\{T\}$ instead of $T$ since the $\mathcal{T}$-ground event is a singleton subset of $\Theta_i$.

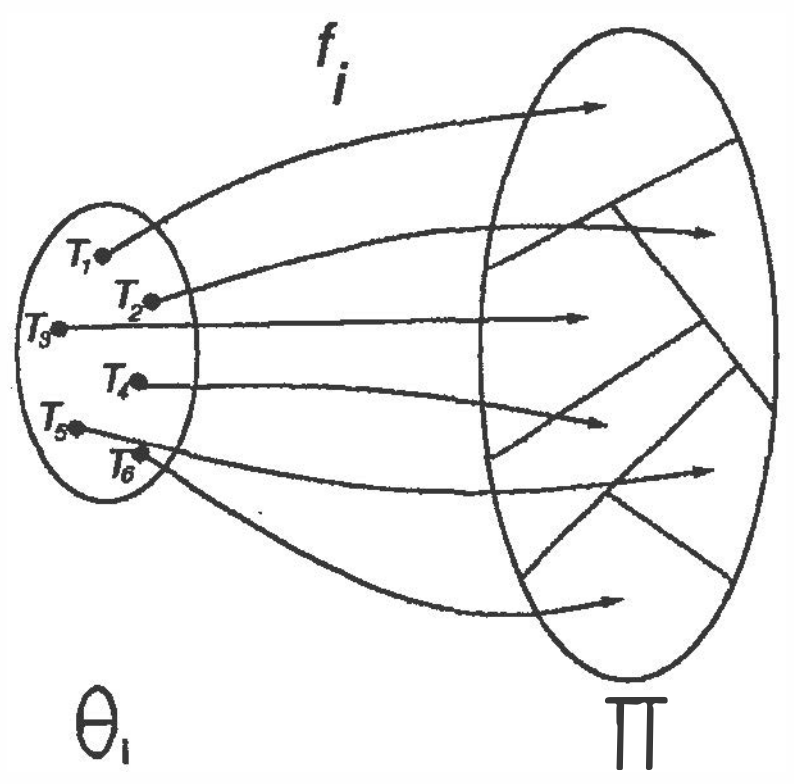

Figure 3: The mapping $f_i$ forms a direct correspondence of elements of $\Theta_i$ to subsets of $\Pi$, and hence a partition of $\Pi$.

The relationship between $\Theta_i$ and $\Pi$ resembles the *refinement* relationship defined by Shafer [21]. In fact, the mapping $f_i$ is very similar to what Shafer terms a *refining mapping*; however, this should not be confused with our use of *refinement* in the context of creating new approximations for $\mathcal{T}_i$.

## 4.2 Compact Representation of Events on $\mathcal{S}$, $\mathcal{S}$-Refinements, and $\mathcal{S}$-covers

We first describe the type of refinements and covers that will be introduced, and then formally present the details. We are interested in performing refinements on $\mathcal{S}$ in a manner similar to that of $\mathcal{T}$-refinement. Consider a $\mathcal{T}$-cover, $C_i$ on some $\mathcal{T}_i$. A partition of $\Pi$ is defined through the application of $f_i$, which we can consider as an $\mathcal{S}$-cover. Any $\mathcal{T}$-refinements that were performed on $\mathcal{T}_i$ could also be considered as $\mathcal{S}$-refinements, yielding $\mathcal{S}$-covers. If we select some $\mathcal{T}$-ground event, say $\{T\}$ from $C_i$, let $A$ denote the corresponding $\mathcal{S}$ event. All segmentations in $A$ contain $T_i$ as a segment; hence, the only variations allowed within $A$ are different groupings of the regions in $\mathcal{R} - T_i$ into segments. The selection of a new initial region $R_j \in \mathcal{R} - T_i$ can be used to form a $\mathcal{T}_j$ that only considers region groupings from $\mathcal{R} - T_i$. A $\mathcal{T}$-cover on $\mathcal{T}_j$ can be considered as a partition of $A$, which in turn represents an $\mathcal{S}$-cover (while holding the remaining events in $C_i$ fixed) that is a closer approximation than the one that contained $A$. A $\mathcal{T}$-ground event, $T_j \in \mathcal{T}_j$ maps to the $\mathcal{S}$ event corresponding to the set of all segmentations that contain *both* $T_i$ and $T_j$. The set $\Pi$ can be further partitioned through the consideration of a new initial region from $\mathcal{R} - T_i - T_j$, and the process continues until $\mathcal{S}$ events are obtained that represent individual segmentations. This description provides the basic strategy, which will next be formalized.

Any event on $\mathcal{S}$ constructed in the manner just discussed can be implicitly represented by a set of segments, $F$, an include set, $I$, and an exclude set, $E$. The elements of $F$ are the segments obtained in a sequence of $\mathcal{T}_i$ constructions. The sets $I$ and $E$ are the include and exclude sets of $\tau(I,E)$, an event in the current $\mathcal{T}_i$ construction. We will use $(\Theta_i, \mathcal{B}_i, P)$ to denote the current $\mathcal{T}_i$. Formally, we represent an event on $\mathcal{S}$ by a function $\sigma$, in which

$$\sigma(F,I,E) = \{S \in \Pi : T \in S \; \forall T \in F\} \bigcap f_i(\tau(I,E)). \quad (7)$$

As defined in Section 4.1, $f_i$ is the function that maps events in $\mathcal{B}_i$ to their corresponding events on $\mathcal{S}$. Thus, (7) represents the set of segmentations that include every segment in $F$ and some segment from $\tau(I,E)$.

An $\mathcal{S}$-*cover*, $\mathbf{C}$, is a set of pairwise-disjoint events in $\mathcal{A}$ that form a partition of $\Pi$. As with $\mathcal{T}$-covers, there is a partial ordering on $\mathcal{S}$-covers; and, as with $\mathcal{T}$-covers, it is possible to construct a finer $\mathcal{S}$-cover from an existing $\mathcal{S}$-cover by performing an $\mathcal{S}$-*refinement operation*. An $\mathcal{S}$-refinement cannot be performed on a singleton element of an $\mathcal{S}$-cover, which is termed $\mathcal{S}$-ground event. As a $\mathcal{T}$-ground event refers to a single segment, a $\mathcal{S}$-ground event refers to a single segmentation.

An $\mathcal{S}$-refinement is performed by partitioning the $\mathcal{S}$-*refinement event*, $A_\rho = \sigma(F_\rho, I_\rho, E_\rho)$, into two finer events, $A_I$ and $A_E$. This is achieved by applying the $\mathcal{T}$-refinement operation to $\tau(I_\rho, E_\rho)$, for a $\mathcal{T}$-refinement region $R_\rho$. For the case of $\mathcal{S}$-refinement, we will refer to $R_\rho$ as the $\mathcal{S}$-*refinement region*. We impose the constraint that $R_\rho$ be adjacent to some region in $I_\rho$, and that it not be in any of $F_\rho$, $I_\rho$, or $E_\rho$. In the case in which $\tau(I_\rho, E_\rho)$ is a not a $\mathcal{T}$-ground event, the two $\mathcal{S}$-refined events are

$$A_I = \sigma(F_\rho, I_\rho \cup \{R_\rho\}, E_\rho) \quad (8)$$

and

$$A_E = \sigma(F_\rho, I_\rho, E_\rho \cup \{R_\rho\}). \quad (9)$$

**Proposition 1** *If $A_\rho = \sigma(F_\rho, I_\rho, E_\rho)$, and $\tau(I_\rho, E_\rho)$ is not a $\mathcal{T}$-ground event, then $A_I$ and $A_E$, given above, form a disjoint partition of $A_\rho$.*

It is possible that $\tau(I_\rho, E_\rho)$ may be a $\mathcal{T}$-ground event. In this case, the construction of a new $\mathcal{T}$ must be initiated. We select some region, $R_j$, that is not in any of $F_\rho$, $I_\rho$, or $E_\rho$ as the initial region for the new $\mathcal{T}$. It is convenient to use an equivalent representation for $\sigma(F_\rho, I_\rho, E_\rho)$, as given in the following proposition.

**Proposition 2** *For an event on $\mathcal{S}$, $\sigma(F_\rho, I_\rho, E_\rho)$, in which $\tau(I_\rho, E_\rho)$ is a $\mathcal{T}$-ground event on $\mathcal{T}_i$, and for some region $R_j$ not in any of $F_\rho$, $I_\rho$, or $E_\rho$*

$$\sigma(F_\rho, I_\rho, E_\rho) = \sigma(F_\rho \cup \{I_\rho\}, \{R_j\}, \emptyset). \quad (10)$$

The proposition above simply gives an alternative representation of the $\mathcal{S}$-refinement event that explicitly uses the new initial region, $R_j$. Therefore we can use Proposition 2, when $\tau(I_\rho, E_\rho)$ is a $\mathcal{T}$-ground event on $\mathcal{T}_i$, and the $\mathcal{S}$-refined events correspond to the first $\mathcal{T}$-refinement that is performed on the new space, $\mathcal{T}_j$:

$$A_I = \sigma(F_\rho \cup \{I_\rho\}, \{R_j, R_\rho\}, \emptyset) \quad (11)$$

and

$$A_E = \sigma(F_\rho \cup \{I_\rho\}, \{R_j\}, \{R_\rho\}). \quad (12)$$

Again, these represent a disjoint partition of $A_\rho$.

## 4.3 Incrementally Determining the $\mathcal{S}$ Probability Map

In this section we describe how the $\mathcal{S}$ probability map is expressed in terms of the $\mathcal{T}$ probability maps, which allows the incremental computation of $\mathcal{S}$ probabilities after performing each $\mathcal{S}$-refinement. The incremental determination of the $\mathcal{T}$ probability maps from statistical image models, on which these results depend, is presented in detail in Section 5.

To avoid confusion in this section we will use $P_\Pi$ to denote the probability map on $\mathcal{S}$, and $P_\Theta$ to denote the probability map on $\mathcal{T}_i$. Explicitly, the probability assigned to a $\mathcal{T}$-ground segment event is assigned directly to the corresponding event on $\mathcal{S}$:

$$P_\Pi(f_i(B)) = \sum_{T \in B} P_\Theta(\{T\}) = P_\Theta(B). \tag{13}$$

For any two ground events $T_1 \in \Theta_i$ and $T_2 \in \Theta_j$, the probability the $\mathcal{S}$ event corresponding to the segmentations that contain both $T_1$ and $T_2$ is given by

$$P_\Pi(f_1(\{T_1\}) \cap f_2(\{T_2\})) = P_\Pi(f_2(\{T_2\}) | f_1(\{T_1\})) P_\Pi(f_1(\{T_1\})). \tag{14}$$

When $T_1 \cap T_2 = \emptyset$ we assume that

$$P_\Pi(f_1(\{T_1\}) \cap f_2(\{T_2\})) = P_\Pi(f_2(\{T_2\})) P_\Pi(f_1(\{T_1\})). \tag{15}$$

Using one segment from each of $n$ $\mathcal{T}$'s, we can develop a general expression similar to (15). Cases in which the probability maps on individual $\mathcal{T}$'s are not independent are addressed in [12].

Hence for an $\mathcal{S}$ event $\sigma(F_\rho, I_\rho, E_\rho)$, if $\tau(I_\rho, E_\rho)$ is a ground event, we can determine the probability using Proposition 2. Otherwise, when $\tau(I_\rho, E_\rho)$ is a not a $\mathcal{T}$ ground event, an $\mathcal{S}$ probability is computed by

$$P_\Pi(\sigma(F_\rho, I_\rho \cup \{R_\rho\}, E_\rho)) = P_\Theta(\tau(I_\rho \cup \{R_\rho\}, E_\rho)) \prod_{T \in F_\rho} P_\Theta(\{T\}). \tag{16}$$

Each $P_\Theta(\{T\})$ represents the probability of the ground event $\{T\}$ in its corresponding $\mathcal{T}$.

# 5 REGION-LEVEL UNCERTAINTY

The purpose of this section is to present expressions that can be used to make probability assignments at refinement steps when building a $\mathcal{T}$ or $\mathcal{S}$ representation. We develop a general statistical context which formulates uncertainty at the data-level. This context pertains to image models that are often used in computer vision. Through the use of Proposition 3, the data-level uncertainty is related to a probabilistic assessment of region-level uncertainty, needed for refinement.

One primary issue must be considered: we are not given a complete representation of $P$ on $\mathcal{T}$. This would require one probability assignment for every ground event. If this was given, then the probability of some other event $B$ is simply the summation over all of the ground events that are subsets of $B$.

Recall that each refinement removes one event in a cover of $\Theta_i$ and replaces it with two disjoint events whose union is the original event. The basic strategy in building a $\mathcal{T}$ representation is to determine probability assignments of the new events when this step is performed. This requires deciding how to divide the probability of the original event between the two new events.

There are two basic mechanisms that exert influence on this probability assignment. As we will discuss in Section 6, there is some prior distribution on the sample space. After the application of evidence, some posterior distribution is obtained. Before constructing the representation, a prior distribution will be defined implicitly on $\mathcal{T}$. Model-based evidence will be used, along with the prior distribution, to determine probability assignments at the refinement step.

Using the refinement mapping, successive partitions are constructed from $\Theta_i$ as prescribed by (3). Recall that in this operation, after selecting $B_\rho$ and $R_\rho$, we partition $B_\rho = \tau(I_\rho, E_\rho)$ into $\tau(I_\rho \cup \{R_\rho\}, E_\rho)$ and $\tau(I_\rho, E_\rho \cup \{R_\rho\})$. For probabilistic consistency, it is necessary to have

$$P(\tau(I_\rho, E_\rho)) = P(\tau(I_\rho \cup \{R_\rho\}, E_\rho)) + P(\tau(I_\rho, E_\rho \cup \{R_\rho\})). \tag{17}$$

It is assumed inductively that $P(B_\rho)$ is known, and that the two probabilities on the right side of (17) must be determined. Before the first refinement is performed, $B_\rho = \Theta_i$, and $P(\Theta_i) = 1$, reflecting the starting condition. At each iteration we will have $P(\tau(I_\rho, E_\rho))$ and need to determine probability assignments on the right side of (17) while making use of priors and model-based evidence. This will determine the probability assignments at each refinment, and hence the probabilities for all of the events in a cover.

It has been assumed in (17) that the probability of $B_\rho$ is never altered by the refinement operation. In general, it could be the case that evidence about $R_\rho$ could cause $P(B_\rho)$ to increase or decrease. Although in this paper we do not describe specific models of homogeneity, it is important to note that in the most general setting a model could be considered that causes $P(B_\rho)$ to change after $R_\rho$ is considered. This is precisely the issue that arises with taxonomic hierarchies, analyzed by Pearl [16]. An efficient method of propagating evidence-based, posterior probabilities throughout a hierarchy of events is presented in Pearl's work, but the construction of the hierarchy by the refinement mapping is not considered. Models that cause $P(B_\rho)$ to change are much more difficult to analyze in our context.

Note that since $P(\tau(I_\rho, E_\rho))$ is known inductively we only need to determine one of the terms on the right side of (17). An alternative way to represent the first of these terms is by

$$P(\tau(I_\rho \cup \{R_\rho\}, E_\rho)) = P_I \; P(B_\rho) \tag{18}$$

in which $P_I$ is called the *membership probability*, and as shown in [12] can be expressed as

$$P_I = P(\tau(\{R_i, R_\rho\}, \emptyset) \,|\, \tau(I_\rho, E_\rho)). \tag{19}$$

Equation (19) is expressed in a form explicitly indicating the importance of adding $R_\rho$ to $I_\rho$ or $E_\rho$. This is the fundamental distinction between the event $B_\rho$ and the two refined events. It is natural to expect that the probability due to evidence will depend directly on the new region that has been brought into consideration, and this has been precisely represented by these expressions.

## 5.1 The Posterior Evidence-Based Membership Probability

In this section we present expressions for the membership probability corresponding to the first $\mathcal{T}$-refinement, considering only the homogeneity of $R_i \cup R_\rho$. The expressions resulting when $E_\rho$ and $I_\rho$ contain regions (besides $R_i \in I_\rho$) are straightforward to derive, and are presented in [12]. We have found experimentally that in most cases there is a negligible difference in probabilities when the information in the other regions in $I_\rho$ and $E_\rho$ is added.

With every image element, $\mathbf{x}$, we associate a random vector $\mathbf{X}$, representing the image information, which may be 3D position, intensity, color, or other information. For each $R_k \in \mathcal{R}$ we define the following four components [13], similar to those used previously in MRF contexts:

- **Parameter space:** A random vector, $\mathbf{U}_k$, which could, for instance, represent a space of polynomial surfaces.
- **Observation space:** A random vector, $\mathbf{Y}_k$, obtained as a function of the data $\mathbf{x} \in R_k$.
- **Degradation model:** A conditional density, $p(\mathbf{y}_k|\mathbf{u}_k)$, which models noise and uncertainty.
- **Prior Model:** An initial parameter space density, $p(\mathbf{u}_k)$.

We have shown that for two regions, $R_i$ and $R_\rho$, the posterior probability that $R_i \cup R_\rho$ is homogeneous, given a prior probability, $P_0$, is determined through the following proposition [12]:

**Proposition 3** *Given the observations* $\mathbf{y}_i$ *and* $\mathbf{y}_\rho$, *the posterior membership probability is*

$$P(\tau(\{R_i, R_\rho\}, \emptyset)|\mathbf{y}_i, \mathbf{y}_\rho) = \frac{1}{1 + \lambda_0\, \lambda_1(\mathbf{y}_i, \mathbf{y}_\rho)}, \quad (20)$$

*in which*

$$\lambda_0 = \frac{1 - P_0}{P_0} \quad (21)$$

*and* $\lambda_1(\mathbf{y}_i, \mathbf{y}_\rho) =$

$$\frac{\left[\int p(\mathbf{y}_i|\mathbf{u}_i)p(\mathbf{u}_i)d\mathbf{u}_i\right]\left[\int p(\mathbf{y}_\rho|\mathbf{u}_\rho)p(\mathbf{u}_\rho)d\mathbf{u}_\rho\right]}{\int p(\mathbf{y}_i|\mathbf{u}_{i\rho})p(\mathbf{y}_\rho|\mathbf{u}_{i\rho})p(\mathbf{u}_{i\rho})d\mathbf{u}_{i\rho}}. \quad (22)$$

Above, $\mathbf{u}_{i\rho}$ represents the parameter space corresponding to $R_i \cup R_\rho$. The $\lambda_0$ and $\lambda_1(\mathbf{y}_i, \mathbf{y}_\rho)$ ratios represent an interesting decomposition into prior and posterior factors.

We next briefly present expressions that pertain to the use of multiple independent models. If we have $m$ independent observations spaces and parameter spaces, we express the posterior membership probability as

$$P(\tau(\{R_i, R_\rho\}, \emptyset)|\mathbf{y}_i^1, \ldots, \mathbf{y}_i^m, \mathbf{y}_\rho^1, \ldots, \mathbf{y}_\rho^m). \quad (23)$$

This membership probability becomes

$$\frac{1}{1 + \lambda_0 \prod_{l=1}^{m} \lambda_k(\mathbf{y}_i^l, \mathbf{y}_\rho^l)} \quad (24)$$

in which $\lambda_k(\mathbf{y}_i^l, \mathbf{y}_\rho^l)$ is similar to (22).

## 6 OBTAINING PRIORS

Bayesian approaches require the specification of prior distributions. The general goal is to reflect some kind of uniformity, due to the lack of information that affects the probability distribution. It might be the case that one wants to introduce some bias through the priors, but this discussion will primarily be concerned with trying to eliminate unwanted bias to yield uniformity. In this section, we describe three possible specifications of prior distributions on $\mathcal{T}$ and $\mathcal{S}$ probability spaces. Each of these specifications corresponds to a particular definition of uniformity over $\mathcal{T}$ or $\mathcal{S}$. In Section 5, we discussed the how the image models are applied to yield a posterior probability distribution.

The first kind of prior uniformity, termed ***segmentation uniformity***, is the condition that all segmentations have equal prior probability, i.e.,

$$P(\{S\}) = \frac{1}{|\Pi|} \ \forall S \in \Pi, \quad (25)$$

in which $|\Pi|$ is the number of possible segmentations. This appears to be the most natural definition of uniformity. The difficulty with segmentation uniformity is that it requires enumerating $\Pi$ before being able to determine the prior. The methods that have been discussed are aimed at avoiding this enumeration. Hence, segmentation uniformity is difficult to explicitly use; however, it serves as a reference for comparing other types of uniformity.

The second kind of prior uniformity, which will be called *segment uniformity*, specifies that each segment in $\mathcal{T}$ has equal prior probability. Specifically, for a space $\Theta_i$

$$P(\{T\}) = \frac{1}{|\Theta_i|} \ \forall T \in \Theta_i. \quad (26)$$

Segment uniformity appears to be a natural choice; however, segment uniformity does not imply segmentation uniformity, except for the special case in which $f(\{T\})$ contains the same number of elements, for all $T \in \Theta_i$. This is implied by the probability constraint (13). Thus, in general, with respect to segmentation uniformity, segment uniformity can be considered as a kind of bias.

The third and final type of uniformity that we consider is *membership uniformity*. Membership uniformity reflects the assumption that for any $\mathcal{T}$-refinement, the prior probabilities associated with the two $\mathcal{T}$-refined events are equal. This corresponds to the assumption that the prior probability of including a region in a segment is equal to the prior probability of excluding that region from the segment. Membership uniformity does not imply segment uniformity, except in the special case when $I \cup E$ has the same number of regions for all $\mathcal{T}$-ground events, $\tau(I, E) \in \mathcal{T}$.

Our experiments indicate that the bias due to priors is readily overcome when evidence is strong. We have also observed that membership uniformity is usually closer to segmentation uniformity than it is to segment uniformity. This is due to the fact that segments with fewer regions (given higher prior probability) tend to cause more $\mathcal{T}$'s to be constructed than larger segments. The probabilities on $\mathcal{S}$ are obtained from these individual $\mathcal{T}$'s using (16). As the number of segments grows, the prior probability tends to decrease, compensating for the small-segment bias with respect to segment uniformity.

## 7 ALGORITHMS

We have developed efficient algorithms that construct $\mathcal{T}$ and $\mathcal{S}$ representations from image data. For each space we have an algorithm that represents the $n$ ground events that have highest probability. They iteratively utilize the refinement operation, applied at each iteration to the non-ground event that has highest probability. Hence, we choose $B_\rho$ to maximize the probability that it contains ground events that have high probability in the space. We also choose $R_\rho$ to maximize the efficiency of the representation. Finally, the algorithms have a simple termination criterion that guarantees the best $n$ segments or segmentations are represented (as ground events) well before enumerating all of them. Further details appear in [12].

# 8 AN EXPERIMENTAL EXAMPLE

To illustrate the theory presented in this paper in an application, we present one example in which 20 best segmentations are represented for a given range image[5]. A synthetic image which consists of 10000 data points (100 x 100) is shown in 4.(a). When the points are projected into the $xy$ plane, there is integer spacing between adjacent points. In the $z$ direction, there is one four-sided pyramid in the image, with a plane in the background. Note that the height of the pyramid is distorted in the figure: the $x_1$ and $x_2$ coordinates range from 0 to 100, while the height of the pyramid, given by maximum value of $x_3$, is only 12. This makes the problem more challenging than the figure may suggest. There is zero-mean Gaussian noise with $\sigma^2 = 0.1$. Figure 4.(b) shows the set of regions, $\mathcal{R}$, that was presented to the algorithm.

The parameter space for this example represents a set of possible planes. The observation space is the sum-of-distance-squared error from the data points to a plane given by a particular parameter space value. Regions that would lie in the same plane without noise are together homogeneous. Using this model, Figure 5 shows the resulting top segmentations.

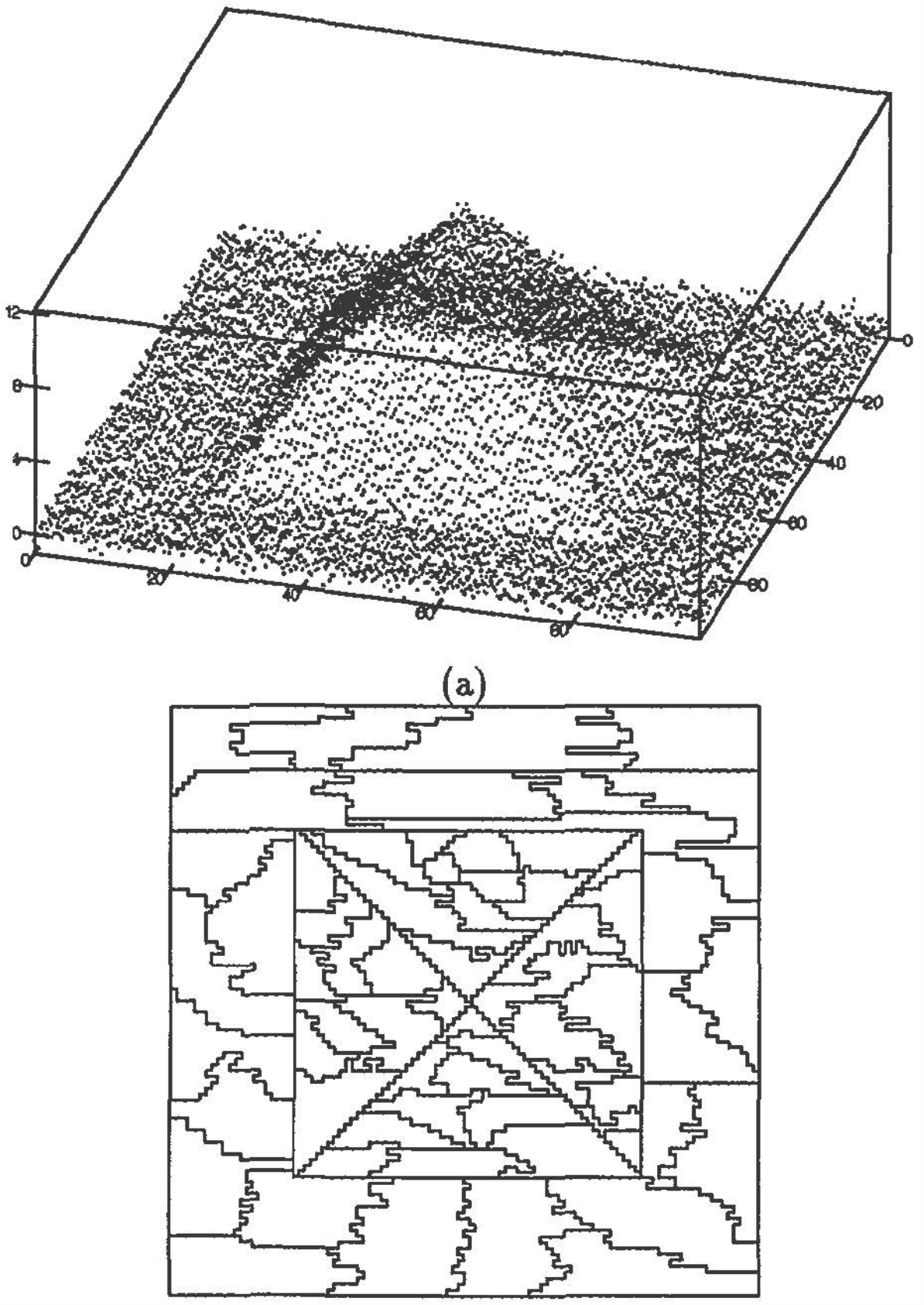

(b)

Figure 4: (a) The data with $\sigma^2 = 0.1$; (b) the set of regions, $\mathcal{R}$

# 9 CONCLUSION

To judge the effectiveness of our approach, we have executed many experiments on both real and synthetic range data. The synthetic data has allowed controlled experiments to be performed in which we could vary the amount of noise and other parameters to see the effects on the resulting probability distributions. As we expect, as the noise level increases the posterior probabilities tend closer to the priors. Results from the real range data indicate that we developed a theoretical and computational approach that can cope with the difficulties imposed by real images.

We have applied the general model presented in Section 5.1 to the class of implicit polynomial surfaces. The resulting high dimensional parameter spaces required us to use Monte-Carlo integration techniques to evaluate the membership probability. We can also presently obtain segmentations using MRF texture models or parametric polynomial models on intensity data as parameter spaces.

## Acknowledgements

This work was sponsored by NSF under grant #IRI-9110270. We are grateful for the comments provided by the anonymous reviewers.

[5]A range image is a collection of points in $\Re^3$, stored in a 2D array.

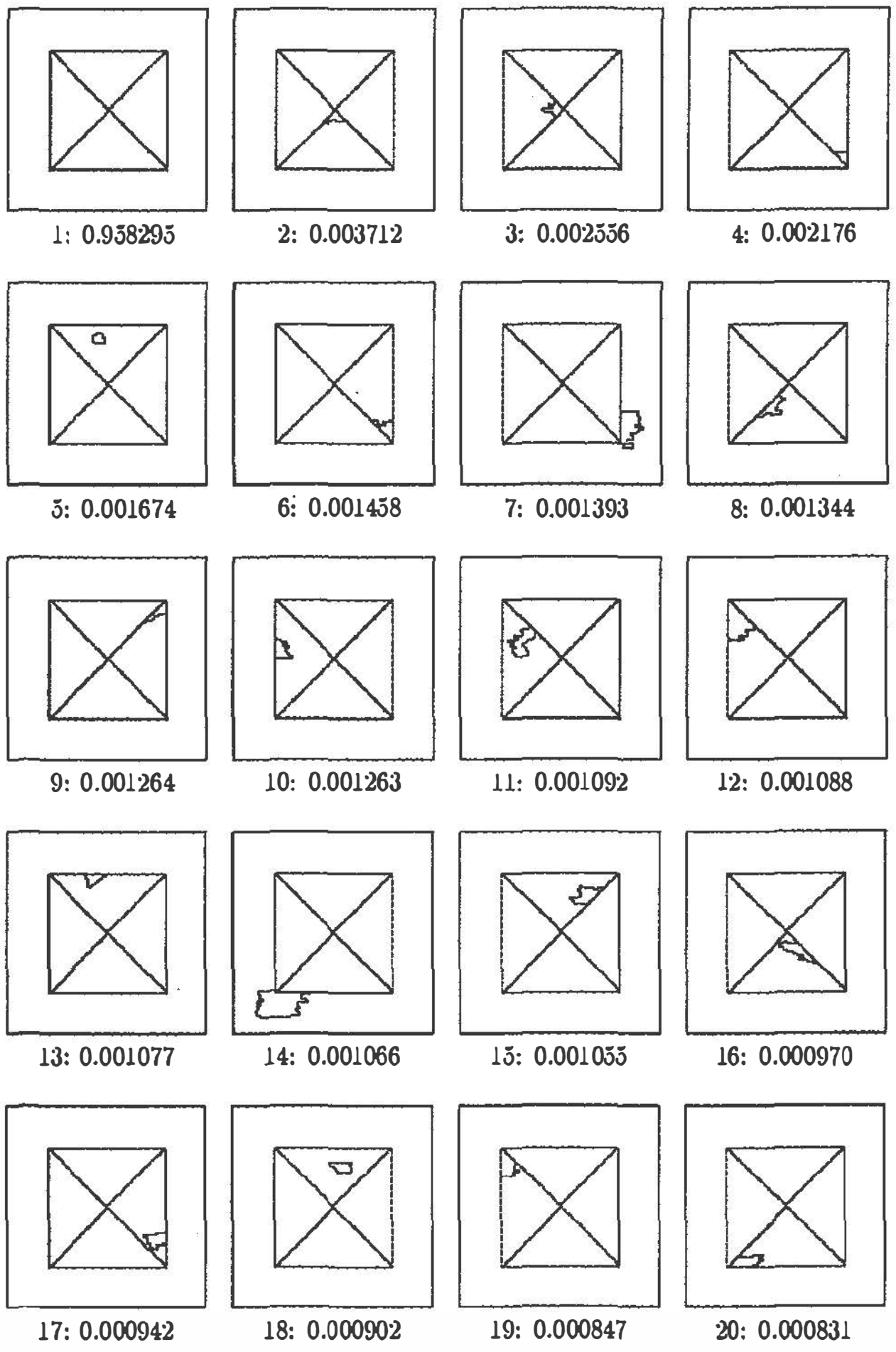

Figure 5: Twenty segmentations that have highest probability in $\Pi$. In this experiment, $\sigma^2 = 0.1$, and the $IE$-independent model is in use